\documentclass[smallcondensed]{svjour3}
\usepackage{graphicx}

\usepackage{times}
\usepackage{amsmath,bm}
\usepackage{xfrac}
\usepackage{paralist}
\usepackage{algorithmic}
\usepackage{amsfonts}
\usepackage[inoutnumbered,linesnumbered,ruled,vlined]{algorithm2e}
\usepackage{color}
\usepackage{array}
\usepackage{multicol}
\usepackage{multirow}
\usepackage{rotating}
\usepackage{tablefootnote}
\usepackage{subcaption}
\usepackage{footmisc}
\usepackage{xcolor}
\usepackage[misc,geometry]{ifsym} 
\usepackage{latexsym}
\usepackage{colortbl}
\usepackage{booktabs}
\usepackage{hyperref}
\usepackage{balance}
\usepackage{xspace}
\usepackage{enumerate}
\usepackage{enumitem}

\newcommand{\gyx}[1]{{\color{black}#1}}
\newcommand{\rv}[1]{{\color{black}#1}}
\newcommand{\bertsubs}{\textsf{BERTSubs}\xspace}
\newcommand{\bertmap}{\textsf{BERTMap}\xspace}

\newcommand{\owlvec}{$\text{OWL2Vec}^{*}$\xspace}
\newcommand{\sroiq}{\ensuremath{\mathcal{SROIQ}}\xspace}

\begin{document}
%
\title{
Contextual Semantic Embeddings for Ontology Subsumption Prediction 
%
}

%
%

\author{Jiaoyan Chen  \and Yuan He \and Yuxia Geng \and Ernesto~Jim\'{e}nez-Ruiz \and Hang Dong \and Ian Horrocks 
}


\institute{Jiaoyan Chen \at Department of Computer Science, The University of Manchester, UK. \email{jiaoyan.chen@manchester.ac.uk}
\and
Yuan He, Hang Dong, Ian Horrocks \at Department of Computer Science, University of Oxford, UK.
              \email{(yuan.he, hang.dong, ian.horrocks)@cs.ox.ac.uk}           
           \and
           Yuxia Geng \at
              College of Computer Science and Technology,  Zhejiang University, China. 
              \email{gengyx@zju.edu.cn}
           \and
           Ernesto Jim\'{e}nez-Ruiz \at City, University of London, UK and
              Department of Informatics,  University of Oslo, Norway.
             \email{ernesto.jimenez-ruiz@city.ac.uk}
}

\date{Received: date / Accepted: date}

\maketitle              
\begin{abstract}
Automating ontology construction and curation is an important but challenging task in knowledge engineering and artificial intelligence. 
Prediction by machine learning techniques such as contextual semantic embedding is a promising direction, but the relevant research is still preliminary especially for expressive ontologies in Web Ontology Language (OWL).
In this paper, we present a new subsumption prediction method named \bertsubs for classes of OWL ontology.
It exploits the pre-trained language model BERT to compute contextual embeddings of a class, where customized templates are proposed to incorporate the class context (e.g., neighbouring classes) and the logical existential restriction.
\rv{\bertsubs is able to predict multiple kinds of subsumers including named classes from the same ontology or another ontology, and existential restrictions from the same ontology.}
Extensive evaluation on five real-world ontologies for three different subsumption tasks has shown the effectiveness of the templates and that \bertsubs can dramatically outperform the baselines that use (literal-aware) knowledge graph embeddings, non-contextual word embeddings and the state-of-the-art OWL ontology embeddings.

\end{abstract}
\section{Introduction}

Ontologies, which often encompass categorizations as well as naming and definition of concepts (classes), properties, logical relationships and so on, are widely used to represent, manage and exchange general or domain knowledge.
They are sometimes regarded as a kind of knowledge graphs (KGs) or act as the schema parts of KGs.
A large number of ontologies have been developed for many domains such as the Semantic Web, bioinformatics, health care, geography and so on \cite{staab2010handbook}, and continue to grow in both science and industry.

High quality ontologies, however, are still created and maintained mainly by human curators.
Many curation tasks cost an extreme large amount of labour, and some of them even cannot be manually addressed as the ontology scale grows. 
Supporting tools, which are expected to assist curators and/or automate some curation procedures, thus become urgently needed \cite{horrocks2020tool}. 
Since an ontology's backbone is usually a set of hierarchical concepts for representing taxonomies, curating concepts and their hierarchy (i.e., a set of subsumption relationships) become especially important.
Relevant tasks such as inserting new concepts, completing subsumptions and matching concepts across ontologies are quite common but very challenging due to difficulties in e.g., capturing the concept meaning.

The existing ontology management tools such as Prot{\'e}g{\'e} mainly provide some interfaces for manual ontology operations \cite{musen2015protege}.
Some ontology visualization methods could help human curators discover some missing subsumption relationships; for example, Ochs et al. \cite{ochs2016unified} extracted an abstract network to summarize the architecture and content of the original concepts.
Such tools and methods, however, cannot automate ontology concept curation or discover subsumptions by themselves. 
Traditional ontology reasoners such as HermiT \cite{glimm2014hermit} and ELK \cite{kazakov2014incredible} exploit logical semantics defined by Web Ontology Language (OWL) \cite{bechhofer2004owl} or Semantic Web Rule Language (SWRL) \cite{horrocks2004swrl} can deductively infer potential subsumptions.
The logical semantics, however, have to be manually defined and are usually far from enough in real-world ontologies for avoiding over expression which will lead to inconsistency.
Thus such symbolic inference cannot discover most plausible subsumptions.

With the development of representation learning, semantic embedding techniques, which represent symbolics such as words and concepts in a sub-symbolic (vector) space with their semantics e.g., relationships concerned, have recently been widely applied in KG curation tasks such as predicting relational facts (a.k.a. KG completion or link prediction) \cite{bordes2013translating,yang2014embedding,lin2015learning,wang2017knowledge}. 
The majority of these studies aim at KGs composed of relational facts, while the research on real-world OWL ontologies, which are often composed of class hierarchies, literals (including text) and logical expressions, is still preliminary.
Some KG embedding methods based on geometric models or Graph Neural Networks could be directly extended to  embed class hierarchies \cite{nickel2017poincare,vilnis2018probabilistic,zhang2020learning} and even partial logical expressions \cite{kulmanov2019embeddings,xiong2022box,garg2019quantum}, but they do not currently consider an ontology's lexically-rich annotations
such as class labels and definitions, all of which contain important complementary semantics and play a critical role in real-world ontologies. For example, the labels of two classes  ``soy milk'' and ``soybean food'' can partially indicate that the two classes have the subsumption relationship according to natural language understanding.

Some very recent ontology embedding studies take such textual semantics into consideration with state-of-the-art performance achieved in concept subsumption prediction tasks;
Chen et al. \cite{chen2021owl2vec} transformed an OWL ontology into sequences by OWL to RDF (Resource Description Framework) projection and random walk, used Word2Vec \cite{mikolov2013efficient} for training concept embeddings, and predicted missing concept subsumption within the ontology; 
while Liu et al. \cite{liu2020concept} developed an ad-hoc strategy to transform concepts and their neighbouring concepts into sequences, fine-tuned a pre-trained language model (PLM) named BERT for concept embeddings \cite{devlin2019bert}, and predicted subsumers of a new isolated concept that is to be inserted into the ontology.
However, concept subsumption prediction for OWL ontologies have not been widely investigated and still has much space to explore, especially on jointly embedding and utilizing a concept's textual annotations and context in the ontology (such as its neighbouring classes). 
Previous general OWL ontology embedding methods such as \owlvec\cite{chen2021owl2vec} and OPA2Vec \cite{smaili2019opa2vec} still use non-contextual word embedding models, which generate one vector for each token no matter where the token appear and have been shown to perform worse than recent Transformer-based contextual word embedding models such as BERT in many sequence learning tasks.
Meanwhile, some complicated but important concept subsumptions, such as the subsumption between a named concept and a complex concept defined in OWL by e.g., existential restriction, and the subsumption between two concepts from different ontologies, have not yet been explored in the current studies.
Liu et al. \cite{liu2020concept} considered new concepts to insert which have quite different context as the (complex) concepts that are already in the ontology.

In this study, we aim to develop a general method that can predict \textit{(i)} the missing concept (class) subsumptions within an ontology and \textit{(ii)} the subsumptions between concepts from two ontologies.
The former involves subsumers of both named concepts and property existential restrictions, and can be applied to complete an ontology; while the latter can be applied to align two ontologies for knowledge integration.
Note that discovering inter-ontology subsumptions is a typical complex ontology alignment task \cite{zhou2020geolink}, and has been rarely investigated, especially with machine learning techniques.
Considering the state-of-the-art performance in many text understanding tasks, we adopt the transformer-based PLM, BERT, for concept embedding, and name our method as \bertsubs.
It models concept subsumption prediction as a downstream classification task for BERT fine-tuning, where three different templates, i.e., Isolated Class (IC), Path Context (PC) and Breadth-based Context (BC), are developed to transform two target concepts and their contexts including the neighbouring concepts and existential restrictions into a pair of sentences as the input.
Note that the templates are general and can be applied to other ontology curation tasks for BERT-based concept semantic embedding.

We extensively evaluated \bertsubs for \textit{(i)} completing  intra-ontology subsumptions of two large-scale real-world ontologies --- the food ontology FoodOn \cite{dooley2018foodon} and the gene ontology GO \cite{gene2008gene}, considering both named class subsumers and existential restriction subsumers, and \textit{(ii)} predicting inter-ontology subsumptions between the health Lifestyles (HeLiS) ontology \cite{dragoni2018helis} and FoodOn, and between the ontologies of
 NCIT (National Cancer Institute Thesaurus) \cite{nicholas2007ncit} and DOID (Human Disease Ontology) \cite{Schriml2018do},
where the ground truth subsumptions extracted from a given set of equivalent class mappings.
\bertsubs often dramatically outperforms the state-of-the-art OWL ontology embeddings methods such as \owlvec \cite{chen2021owl2vec} as well as several other baselines using KG embeddings of TransE \cite{bordes2013translating}, TransR \cite{lin2015learning}, DistMult \cite{yang2014embedding}, HAKE \cite{zhang2020learning}, Text-aware TransE \cite{Schriml2018do}, etc; while the effectiveness of different templates for concept semantic embedding has also been verified. 

The remainder of this paper is organized as follows. 
The next section introduces the preliminaries
including 
the target problem.
Section \ref{sec:methodology} presents the technical details of \bertsubs. 
Section \ref{sec:evaluation} introduces the experiments and the evaluation.
Section \ref{sec:related} gives a comprehensive review of the related works.
The final section concludes the paper with discussion.

\section{Preliminaries}

\subsection{OWL Ontology}

OWL ontologies \cite{DBLP:journals/ws/GrauHMPPS08}
are based on the \sroiq{} description logic (DL) \cite{baader2017introduction}.
An ontology comprises a TBox and an ABox.
The TBox defines atomic concepts and roles, and uses DL constructors such as conjunction (e.g., $C \sqcap D)$, disjunction (e.g., $C \sqcup D$) and existential restriction (e.g., $\exists r.C$) to compose complex concepts, where $C$ and $D$ denote concepts and $r$ denotes a role.
The TBox also includes General Concept Inclusion (GCI) axioms (e.g., $C \sqsubseteq D$) and Role Inclusion axioms 
(e.g., $r \sqsubseteq s$), where $s$ denotes another role.
The ABox is a set of assertions such as concept assertions (e.g., $C(a)$), role assertions (e.g., $r(a, b)$) and individual equality assertions (e.g., $a \equiv b$), where $a$ and $b$ denote two individuals.

\begin{figure*}
\centering
\includegraphics[width=1.0\textwidth]{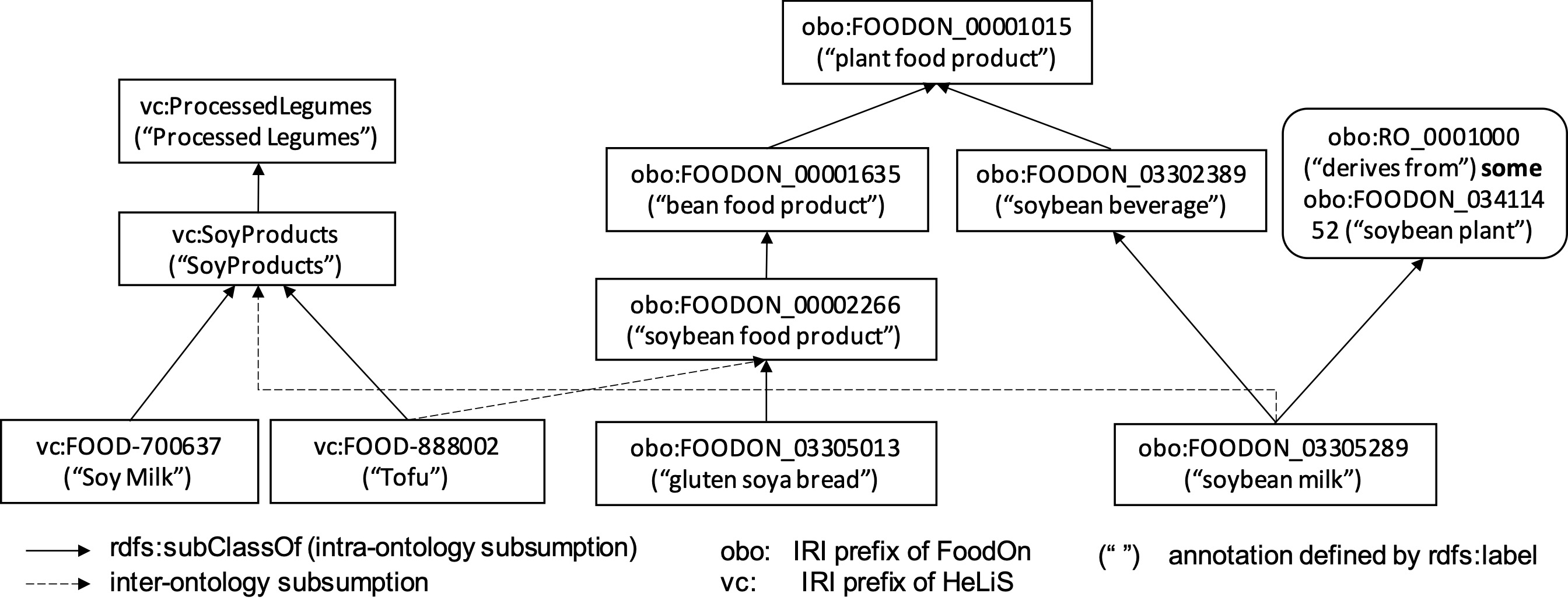}
\vspace{-0.2cm}
\caption{Ontology segments from HeLiS (Left) and FoodOn (Right) with examples of inter-ontology and intra-ontology class subsumptions \label{fig:onto}}
\end{figure*}

In OWL, the aforementioned concept, role and individual are modeled as \textit{class}, \textit{object property} and \textit{instance}, respectively.
An atomic concept corresponds to a \textit{named class}, while the class of a concept by DL constructors is sometimes called \textit{complex class}.
Class, object property and instance can all be referred to as \textit{entity}.
Each entity is uniquely represented by an Internationalized Resource Identifier (IRI).
As shown 
in Figure \ref{fig:onto}, these IRIs may be lexically ``meaningful'' (e.g., \textit{vc:ProcessedLegumes} of HeLiS) or consist of internal IDs that do not carry useful lexical information (e.g., \textit{obo:FOODON\_00002809} of FoodOn); in either case the intended meaning may also be indicated via annotation axioms which can be represented by RDF triples using annotation properties as the predicates; e.g., the class \textit{obo:FOODON\_00002809 (edamame)} is annotated using \textit{rdfs:label} --- a built-in annotation property by RDFS to specify a name string, 
and using \textit{obo:IAO-0000115 (definition)} --- a bespoke annotation property to specify a natural language ``definition''.
To facilitate understanding we will from now on refer to an entity by one of its readable labels.

A GCI axiom $C \sqsubseteq D$ corresponds to a subsumption relation between the class $C$ and the class $D$.
When an ontology is serialised as RDF\footnote{\url{https://www.w3.org/TR/rdf11-concepts/}} triples, each of which is a tuple composed of a subject, a predicate and an object, $C \sqsubseteq D$ is represented as ($C$, \textit{rdfs:subClassOf}, $D$) where \textit{rdfs:subClassOf} is a built-in property in RDFS\footnote{\url{https://www.w3.org/TR/rdf-schema/}}.
For shortness, we abbreviate the subsumption triple as $(C,D)$. 
Real-world ontologies often use a property existential restriction as the subsumer (superclass) of a GCI axiom to (partially) specify the semantics of classes.
For example, in FoodOn \textit{soybean milk} is subsumed by an existential restriction on the property \textit{obo:RO\_0001000 (derives from)} to the class \textit{obo:FOODON\_03411452 (soybean plant)}; this specifies that every instance of \textit{soybean milk} is related to some instance of \textit{soybean plant} via the property \textit{derives from}.
%
Such property existential restrictions are widely used in many real-world ontologies such as FoodOn and SNOMED CT\footnote{\url{https://www.snomed.org/}} due to their relatively high expressivity in representing class semantics and polynomial time complexity in reasoning \cite{baader2005pushing}.
Note that an OWL ontology can support entailment reasoning as implemented by, e.g., HermiT \cite{glimm2014hermit}, through which hidden subsumption axioms can be entailed, e.g., via the transitivity of the subsumption relation and the logical definition of complex classes. 
Entailing class subsumptions using the transitivity of the subsumption relation is sometimes also called inheritance reasoning.
For example, in FoodOn, we can entail that \textit{gluten soya bread} is a subclass of \textit{bean food product} via the intermediate class \textit{soybean food product} and the transitivity of the subsumption relationship.

A class hierarchy, i.e., the pre-order on named classes induced by the subsumption relation (\textit{rdfs:subClassOf}), can usually be extracted from an OWL ontology.
It can simply be the set of all the declared subsumptions between named classes, or computed with entailment reasoning, through which the entailed subsumptions between named classes (except for these by inheritance reasoning) are added.
Sometimes the class hierarchy can be  extended with complex classes and logic expressions such as the property existential restriction.

\subsection{BERT}

\begin{figure*}
\centering
\includegraphics[width=1.0\textwidth]{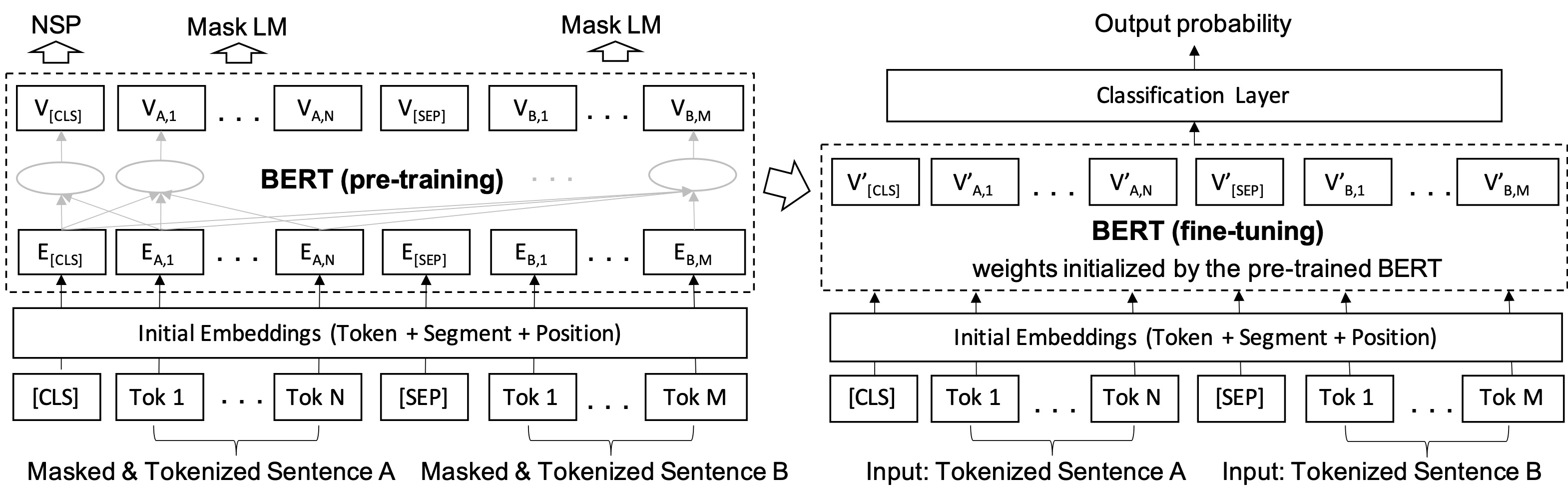}
\vspace{-0.2cm}
\caption{BERT Pre-training (Left) and Fine-tuning for Sentence Pair Classification (Right)} \label{fig:bert}
\end{figure*}

BERT is a contextual word embedding model based on deep bidirectional transformer encoders \cite{devlin2019bert}.
It is often pre-trained with a large general purpose corpus, and then can be applied following a typical fine-tuning paradigm, i.e., it is attached by a classification layer and its parameters are further adjusted towards a specific downstream task with labeled samples, as shown in Figure \ref{fig:bert}.
In pre-training, each input is a sequence composed of a [CLS] special token, two sentences (denoted as A and B), each of which is tokenized e.g., into sub-words by WordPiece \cite{schuster2012japanese}, and a [SEP] special token that separates A and B. 
The embedding of each token is initialized by its one-hot encoding, segment encoding (in A or B) and position encoding.
The parameters of the stacked encoders are learned with two self-supervision tasks: Next Sentence Prediction (NSP) which predicts whether B is following A using the embedding of [CLS], and Masked Language Modeling (LM) which predicts randomly masked tokens in both A and B.

In fine-tuning, the input can be either one sentence with the [CLS] special token or two sentences as in pre-training, depending on the downstream tasks. 
Figure \ref{fig:bert} presents the fine-tuning for the classification of two sentences, where the [CLS] token embedding is fed to a (binary) classification layer for a probabilistic output.
This architecture can support tasks such as predicting whether one sentence (the premise) entails another sentence (the hypothesis) and predicting whether two sentences (phrases) are synonyms, and is adopted by \bertsubs.
Since labeled samples are given in fine-tuning, the model parameters are updated by minimizing a task-specific loss over the given labels.

It is worth mentioning that the pre-trained BERT can also be applied without or with only a little task-specific fine-tuning, by transforming the downstream task into a task adopted in pre-training (e.g., predicting the masked token in a natural language sentence).
Such a paradigm is sometimes known as prompt learning, and its key advantage is relying on no or only a small number of task-specific labeled samples \cite{liu2021pre,gao2021making}.
In our ontology subsumption prediction, an ontology already has quite a large number of given subsumptions that can be used for fine-tuning and it has no high requirement on the computation time as an offline application. 
Thus we prefer the BERT fine-tuning paradigm instead of prompting learning. 

As a contextual word embedding model, BERT and its variants (such as BioBERT which is pre-trained on biomedical corpora \cite{lee2020biobert}) have been widely investigated, with better performance achieved than traditional word embedding and sequence feature learning models such as Recurrent Neural Networks in many Natural Language Processing tasks.
In contrast to non-contextual word embedding models such as Word2Vec \cite{mikolov2013efficient}, which assign each token only one embedding, BERT can distinguish different occurrences of the same token taking their contexts (attentions of surrounding tokens) into consideration. 
Considering the sentence \textit{``the bank robber was seen on the river bank''}, BERT computes different embeddings for the two occurrences of \textit{``bank''} which have different meanings.

\subsection{Problem Statement}

This study aims to predict two kinds of class subsumptions by one general method:
 \begin{itemize}[leftmargin=*]
\item \textbf{Intra-ontology subsumptions}.
Given an OWL ontology $\mathcal{O}$, it is expected to discover plausible class subsumptions so as to complete its class hierarchy.
%
For a named class $C$ in $\mathcal{O}$, the method predicts a score $s$ in $[0,1]$ for each class $D$ in $\mathcal{O}$ to indicate the likelihood that $D$ subsumes $C$; i.e., 
the method ranks the classes in $\mathcal{O}$ according to their likelihood of being a superclass of $C$.
Note $D$ can be either a named class or a property existential restriction.
For example, the named class \textit{soybean food product} and the existential restriction (\textit{derives from} some \textit{seed (anatomical part)}) are expected to be predicted with high scores and ranked in high positions, as subsumers of \textit{soybean milk} (see Figure~\ref{fig:onto}).
In ranking, we exclude the class $C$ itself and its superclasses that can already be logically entailed, such as \textit{plant food product} w.r.t.\ \textit{soybean milk}.
For simplicity, in this paper we call the subsumption between two named classes as \textit{named subsumption} and the subsumption between a named class and a property existential restriction as \textit{existential subsumption}.

\item \textbf{Inter-ontology subsumptions}.
Given two ontologies $\mathcal{O}$ and $\mathcal{O}'$, and a named class $C$ in $\mathcal{O}$,  the method predicts a similar score $s$ as in the first problem, for each named class $D$ in $\mathcal{O}'$ to indicate the likelihood that $D$ subsumes $C$. 
For example, the named class \textit{Soy Product} in HeLiS is expected to be the subsumer of \textit{soybean milk} in FoodOn.
The class pair ($C$, $D$) is regarded as a subsumption mapping between $\mathcal{O}$ and $\mathcal{O}'$.
Such mappings are usually for knowledge integration and do not consider complex classes in $\mathcal{O}'$.

\end{itemize}

\section{Methodology}\label{sec:methodology}

\subsection{Framework}
\begin{figure}
\centering
\includegraphics[width=0.75\textwidth]{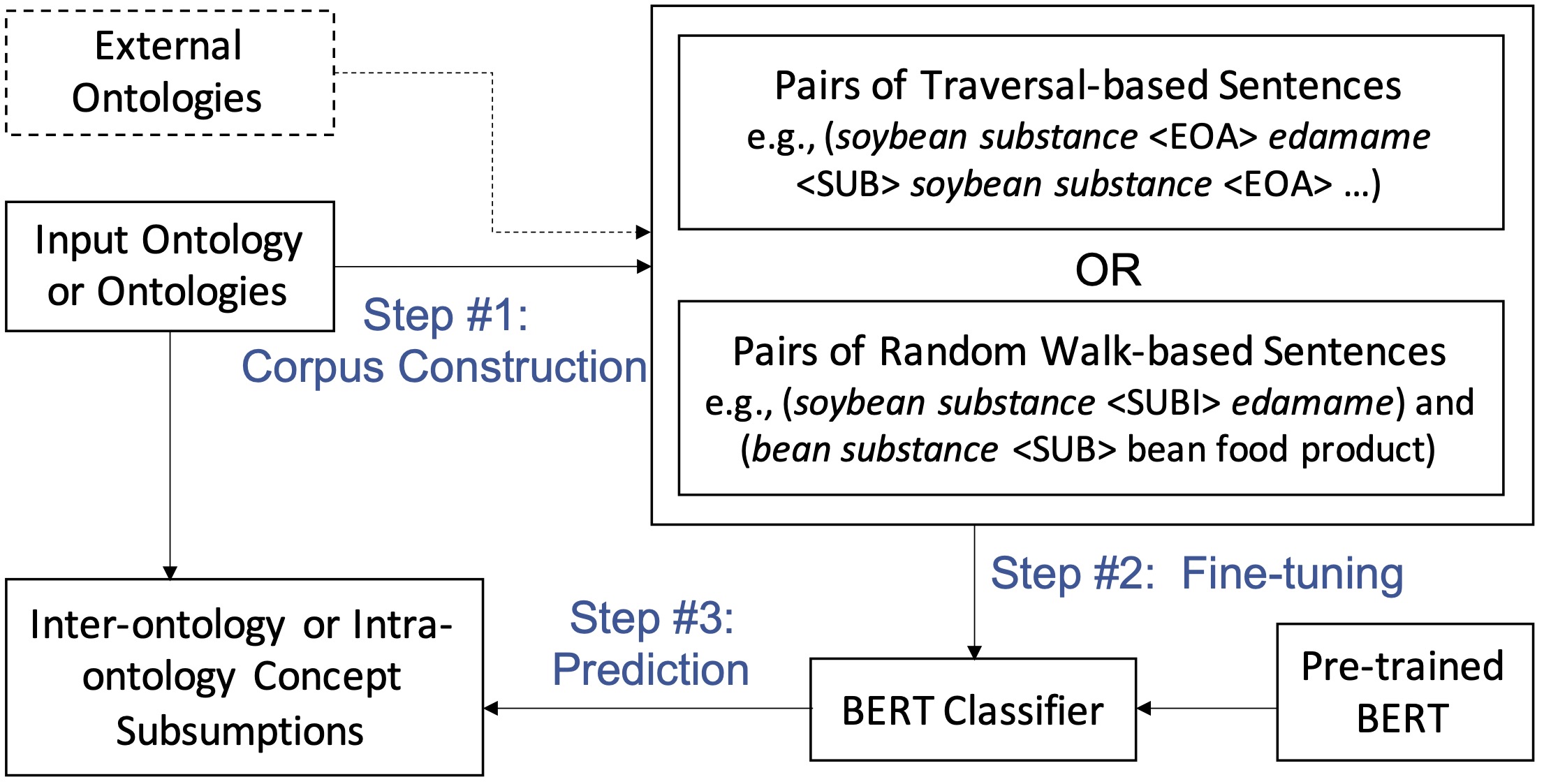}
\vspace{-0.2cm}
\caption{The Framework of \bertsubs} \label{fig:framework}
\end{figure}

Figure \ref{fig:framework} shows the overall framework of \bertsubs. Following a typical BERT fine-tuning paradigm, it mainly includes three parts:
 \begin{itemize}[leftmargin=*]
 \item \textbf{Corpus Construction}. 
 Given an ontology $\mathcal{O}$ (or two ontologies $\mathcal{O}$ and $\mathcal{O}'$), \bertsubs extracts a corpus, i.e., a set of sentence pairs $\mathcal{S}$ by three steps. 
 First, it extracts positive class subsumptions from the class hierarchy of $\mathcal{O}$ (or from the class hierarchies of both $\mathcal{O}$ and $\mathcal{O}'$).
 The current \bertsubs implementation uses the class hierarchy extracted from declared subsumptions without entailment reasoning.
 Second, for each positive subsumption $(c_1, c_2)$, it generates a negative class subsumption by randomly replacing $c_2$ by a named class or existential restriction in the class hierarchy, with all the declared or entailed subsumers of $c_1$ excluded for preventing from generating false negative samples.
 Note the entailed subsumers of $c_1$ are very efficiently computed by inheritance reasoning over the class hierarchy.
 Third, it transforms each subsumption into one or multiple sentence pairs.
It currently has three templates to transform a named class or an existential restriction into a sentence: \textit{Isolated Class}, \textit{Path Context}, and \textit{Breadth-first Class Context}, all of which will be introduced in the remainder of this section.
 Note for predicting named subsumptions, we extract named subsumptions for training, while for predicting existential subsumption, we extract existential subsumptions using the class hierarchy extended with property existential restrictions.
 
 \item \textbf{Model Fine-tuning}. 
 To construct a classifier, a linear layer with dropout is first attached to a pre-trained BERT. It takes as input the embedding of \texttt{[cls]} token from the BERT's last-layer outputs, and transforms the embedding into a 2-dimensional vector. 
 The linear layer is then further attached with a softmax layer to output the score $s$ in $[0,1]$ which indicates the truth degree of the subsumption relationship.
 The parameters of the pre-trained BERT encoders and this new classifier are jointly optimized using the Adam algorithm 
 by minimizing the cross-entropy loss over the samples (labeled sentence pairs). 
 %
Note that the original input sentences, which are sequences of words, are parsed into sub-word (token) sequences by an inherent WordPiece tokenizer \cite{wu2016google} of the pre-trained BERT, before they are input to the BERT encoders.
For example, the original word ``soybean'' is parsed into two tokens ``soy'' and ``\#\#bean''.

 \item \textbf{Prediction}. In prediction, the fine-tuned model predicts a score for each candidate subsumption to test (i.e., a pair of classes) which will be transformed into one or multiple sentence pairs using the same template as used in generating the fine-tuning samples.
When one subsumption is transformed into multiple sentence pairs, they are predicted independently, and the scores are averaged as the score of the subsumption.
 \end{itemize}
 
\subsection{Isolated Class (IC)}\label{sec:isolated}
In this class-to-sentence transformation template, we use the name information of a named class or an existential restriction alone without considering its surrounding classes.
One subsumption $(c_1, c_2)$ will lead to sentences of $\left\{ (s_1, s_2) | s_1 \in L(c_1), s_2 \in L(c_2) \right\}$, where $L(\cdot)$ denotes a class's labels.
For a named class, \bertsubs uses the label defined by \textit{rdfs:label} by default, and can also use the labels defined by some other annotation properties for synonyms, 
such as \textit{obo:hasExactSynonym}\footnote{\textit{obo:} is short for the prefix of \url{http://www.geneontology.org/formats/oboInOwl\#}.} 
and \textit{obo:hasSynonym} in FoodOn. Using multiple labels by different annotation properties lead to many more sentence pairs, and this setting is compared with using one English label by \textit{rdfs:label} in our evaluation (see Section \ref{sec:label}).

For an existential restriction in form of $\exists r.C$ where $C$ is a named class, \bertsubs generates its natural language descriptions as its labels using the template of ``something $L(r)$ \_ $L(C)$''. $L(r)$ and $L(C)$ are the labels of the object property (relation) $r$ and the class $C$, respectively, while \_ is an optional preposition that is added when the property label itself does not end with a preposition, so as to generating natural language text (e.g., ``of'' is added after the FoodOn property \textit{obo:RO\_0000086} (``has quality'')).
As an example, the existential restriction (\textit{obo:RO\_0001000 some obo:FOODON\_03411452}) in Figure \ref{fig:onto} will get the label of ``something derives from soybean plant''.
\bertsubs can also be easily extended to address those existential restrictions in forms of $\exists r.(C_1 \sqcap C_2, ...)$ and $\exists r.(C_1 \sqcup C_2, ...)$, where the restriction class is the conjunction or disjunction of multiple named classes. 
Their templates are ``something $L(r)$ \_ $L(C_1)$ and $L(C_2)$ ...'' and ``something $L(r)$ \_ $L(C_1)$ or $L(C_2)$ ...'', respectively.
For more complicated existential restrictions where the restriction class is a complex class with negation or some other existential restriction (e.g., $\exists r_1.(C_1 \sqcap \exists r_2.C_2)$), the label generation could be quite complicated and we leave it for future extension, where some previous works on generating natural language descriptions from OWL ontology concepts (e.g., \cite{stevens2011automating,kaljurand2007attempto}) could be referred to.

\subsection{Path Context (PC)}\label{sec:depth}
For a class subsumption $(c_1,c_2)$, the PC template first extracts paths starting from $c_1$ and $c_2$, respectively, from the class hierarchies of their corresponding ontologies.
The paths of $c_1$ are generated by depth-first traversal which starts from itself to one of its child classes, continues to its descendants, and stops when it has arrived at a leaf class or the depth has reached the maximum value $d$. 
One path can be denoted as
$p(c_1) = [c_1, c_{1,1}, ..., c_{1,i-1}, c_{1,i}, ..., c_{1,d}]$ 
where $(c_1, c_{1,1})$, $\cdots$, $(c_{1,i-1}, c_{1,i})$ are class subsumptions.
The paths of $c_2$ are also generated by depth-first traversal which starts from itself to one of its parent classes, continues to its ancestors, and stops when it arrives at the top class \textit{owl:Thing} or the depth exceeds the maximum depth of $d$. 
When $c_2$ is a property existential restriction, the depth-first traversal is not applied and the path is $[c_2]$.
When traversing from one class to the next, there could be multiple subclasses or superclasses, and we randomly select at most $w$ of them to reduce the size of paths. $w$ and $d$ are two hyper-parameters.
We extract the path down for $c_1$ and up for $c_2$, because the downside context of $c_1$ has its more fine-grained information, the upside context of $c_2$ has its more general information, and they are more useful for determining whether $c_1$ is a subclass of $c_2$.

With the paths, sentences are then generated by replacing the classes by their labels.
One path of $c_1$, briefly denoted as $p(c_1) = [c_1, c_{1,1} ..., c_{1,d}]$, is transformed into sentences of $$S_p(c_1) = \left\{[s_1 \text{ [SEP] } s_{1,1} \cdots s_{1,d}] | s_1 \in L(c_1), s_{1,1} \in L(c_{1,1}), \cdots,s_{1,d}\in L(c_{1,d}))\right\}$$
where the special token [SEP] separates the labels of every two classes.
The path of $c_2$ is transformed into sentences in the same way, denoted as $S_p(c_2)$. 
For a property existential restriction, its label is generated using the same template as in IC.
Note we have considered introducing a new special token [SUB] instead of using the BERT built-in special token [SEP], for representing the semantics of \textit{rdfs:subClassOf}, but it does not lead to better performance in our evaluation. One potential reason is that it may need a much larger training corpus to let the model learn the new special token's representation.  
With $S_p(c_1)$ and $S_p(c_2)$, the final set of the sentence pairs of the subsumption $(c_1,c_2)$ can be represented as $\left\{ (s_1, s_2) | s_1 \in S_p(c_1), s_2 \in S_p(c_2) \right\}$.
Take the candidate named subsumption (\textit{obo:FOODON\_03302389}, \textit{obo:FOODON\_00002266}) in Figure \ref{fig:onto} as an example, one potential sentence pair with $d$ = $2$ is (``soybean beverage [SEP] soybean milk'', ``soybean food product [SEP] bean food product [SEP] plant food product'').
Since one subsumption could have quite a few sentence pairs due to different paths and different class labels , we set $w$ and $d$ to small numbers (e.g., $4$ and $2$ respectively).

\subsection{Breadth-first Context (BC)}\label{sec:breadth}
Given a class subsumption ($c_1$, $c_2$), the BC template first extracts context class sequences for each class via breadth-first traversal starting from this class over the class hierarchy of its corresponding ontology.
For $c_1$, it first extracts subclasses of $c_1$ for context subsumptions of depth $1$, then extracts subclasses of each subclass achieved at depth $1$ for context subsumptions of depth $2$, and so on.
It stops when the depth has arrived at $d$ or all the subclasses of this depth have been leaf classes. 
$c_1$ and the context subsumptions achieved in order are concatenated to generate the sequence, denoted as  $b(c_1) = [c_1, c_{1,1}, c_1, \cdots$,$c_{1,i}, c_{1,j}, \cdots$], where ($c_{1,1}$, $c_1$), ..., ($c_{1,i}$, $c_{1,j}$) are class subsumptions.
For $c_2$, if it is a named class, its sequence is generated in the same way except that the BC template traverses up for context subsumptions until the depth arrives at $d$ or the superclasses have only the top class \textit{owl:Thing}. 
Its sequence is denoted as $b(c_2) = [c_2, c_{2}, c_{2,1}, \cdots$,$c_{2,i}, c_{2,j}, \cdots$], where ($c_{2}$, $c_{2,1}$), ..., ($c_{2,i}$, $c_{2,j}$) are class subsumptions.
If $c_2$ is an existential restriction, its sequence is simply $[c_2]$ without surrounding classes.
For both $c_1$ and $c_2$, to limit the sequence length, the BC template randomly extracts at most $w$ subclasses at each depth. It also does multiple times of traversal, leading to multiple sentences for $c_1$ and $c_2$.

For each sequence the classes are replaced by their labels to generate sentences.
The sentences of a sequence of $c_1$ are computed as
\begin{align*}
S_b(c_1) =  \big\{
[s_1 \text{ [SEP] } s_{1,1} \text{ [SEP] } s_{1} \text{ [SEP] } ... s_{1,i} \text{ [SEP] } s_{1,j} ...] | \\ s_1 \in L(c_1), s_{1,1} \in L(c_{1,1}), s_{1,i} \in L(c_{1,i}),  s_{1,j} \in L(c_{1,j}) \big\}.
\end{align*}
As in the PC template, we originally considered using a new special token [SUB] to separate two class labels of a subsumption (e.g., $s_{1,i}$ and $s_{1,j}$) but the performance is not as good as simply using the BERT built-in token [SEP] which is to separate two sentences.
Consider \textit{obo:FOODON\_00001015} in Figure \ref{fig:onto} as an example of $c_1$, one of its sentences with depth of $1$ can be ``plant food product [SEP] bean food product [SEP] plant food product [SEP] soybean beverage [SEP] plant food product''.
The sentences of $c_2$, i.e., $S_b(c_2)$, can be calculated in the same way as $c_1$.
The final sentence pairs of $(c_1,c_2)$ are calculated as $\left\{ (s_1, s_2) | s_1 \in S_b(c_1), s_2 \in S_b(c_2) \right\}$.
In evaluation, we also only consider small numbers for $w$ and $d$ to limit the sentence length.
In comparison with PC, the BC template contains more complete contexts of $c_1$ and $c_2$, but also leads to higher redundancy in the sentences.

\section{Evaluation}\label{sec:evaluation}
\subsection{Datasets and Experiment Settings}
\subsubsection{Intra-ontology Subsumption Prediction} 
For experiments on the two intra-ontology subsumption prediction tasks, we used two large-scale real-world OWL ontologies --- the food ontology FoodOn (0.4.8) which represents fine-grained food taxonomies and relevant concepts about e.g., agriculture and plant \cite{dooley2018foodon}, and the gene ontology GO \cite{gene2008gene} of the full version accessed on January 7, 2022\footnote{\url{http://geneontology.org/docs/download-ontology/}}.
Please see Table \ref{tab:datasets1} for more statistics on the two tasks.
Note the concerned existential restrictions refer to those in the class hierarchy in form of $\exists r.C$.
Those more complex existential restrictions in form of e.g., $ \exists r.(C_1 \sqcup C_2)$ and $ \exists r.(\exists r'.C' \sqcap A)$
are currently out of our concern.
For each ontology, the named subsumptions, or the existential subsumptions, are split into $80\%$ for training, $5\%$ for validation, and $15\%$ for testing; the validation and testing subsumptions are masked in training.
We adopt a ranking-based mechanism for evaluating the model performance.
For each validation or testing subsumption $(c_1,c_2)$, we get a set of negative subsumers $C_{neg}$, each of which does not subsume $c_1$, and rank all the classes in $C_{neg} \cup \left\{c_2\right\}$ according to their scores as the subsumers of $c_1$.
According to the ranking position of $c_2$, we calculate the metrics of Mean Reciprocal Rank (MRR), and Hits@$K$ (H@$K$ in short) which means $c_2$ is ranked within top-$K$ positions ($K$ is set to $1$, $5$, $10$). 
For all these metrics, the higher the value, the better the performance.

To distinguish the performance of different models, but dramatically reduce the computation such that complex models and many different settings can be efficiently evaluated, it is expected to consider the most challenging negative subsumers, instead of simply selecting just one random subsumer, or using all the named classes (or existential restrictions) in the class hierarchy. 
If $(c_1,c_2)$ is a named subsumption, we get the negative subsumers $C_{neg}$ by extracting neighbouring named classes of $c_2$ from the class hierarchy.
Specifically, we first extract the one-hop neighbouring named classes of $c_2$, i.e., $C_{neg}^1 = \left\{c | (c,c_2) \text{ or } (c_2,c)\right\}$, and then randomly select at most $m$ seeds from $C_{neg}^1$, and extract the one-hop neighbouring named classes of each seed, merge them as the two-hop neighbouring named classes of $c_2$, denoted as $C_{neg}^2$. We continue until the $h$-hop neighbouring named classes $C_{neg}^k$ have been extracted.
$C_{neg}^1$, $C_{neg}^2$, ..., $C_{neg}^h$ are finally merged, and $50$ negative subsumers are randomly selected from them if the total number exceeds $50$.
Note the subsumptions that can be entailed are not considered in this step for creating $C_2^{n}$, and $m$ and $h$ are set to $8$ and $3$, respectively, for both FoodOn and GO. 
If $(c_1, c_2)$ is an existential subsumption, we get $C_{neg}$ by first randomly selecting $n_1$ from all the concerned property existential restrictions that either have the same property or the same restriction class as $c_2$, then randomly selecting another $n_2$ from the remaining property existential restrictions. 
$n_1$ and $n_2$ are set to $40$ and $10$, respectively, for both FoodOn and GO.
Note we exclude the subsumers of $c_1$ that can be entailed via inheritance reasoning for $C_{neg}$, so as to avoid their impact on the ranking of the ground truth class $c_2$.
On average, FoodOn and GO have $36.2$ and $44.9$ negative subsumers, respectively, for each named subsumption, and have $48.9$ and $50.0$ negative subsumers, respectively, for each existential subsumption.

\begin{table}
\renewcommand\arraystretch{1.3}
\centering
\footnotesize{
\centering
\begin{tabular}{m{1.2cm}<{\centering}|p{1.5cm}<{\centering}|p{2cm}<{\centering}|p{2cm}<{\centering}|p{2cm}<{\centering}}\hline
Ontology & Named Classes & Existential Restrictions & Named Subsumptions & Existential Subsumptions \\ \hline
FoodOn & $28,645$ & $1,187$ & $29,599$ & $6,017$ \\ \hline
GO & $50,757$ & $14,379$ & $70,759$ & $18,833$ \\ \hline
\end{tabular}
\vspace{-0.2cm}
\caption{
Statistics of the two ontologies used for intra-ontology subsumption prediction. }\label{tab:datasets1}
}
\end{table}

For the baselines, we adopt the following kinds:
\begin{itemize}[leftmargin=*]
\item \textbf{Word embedding}. We use the original Word2Vec \cite{mikolov2013efficient} trained by a Wikipedia English article dump in 2018. Each class is represented as the average of the word vectors of its label tokens. The embeddings of the two classes of a subsumption are concatenated and fed to a binary classifier which is trained by the training set. Both Logistic Regression (LR) and Random Forest (RF) are tested as the classifier, and we select the best classifier and its hyper parameters according to the MRR results on the validation set.

\item \textbf{OWL ontology embedding methods}. We test three OWL ontology embedding methods --- \owlvec \cite{chen2021owl2vec}, OPA2Vec \cite{smaili2019opa2vec}, Onto2Vec \cite{smaili2018onto2vec} to embed the classes, and adopt the same binary classifier solution as the word embedding baseline.
\owlvec, OPA2Vec and Onto2Vec can be understood as the ontology corpus trained (or fine-tuned) Word2Vec models.
Specially, for \owlvec, we consider not only a class's word vector (the average of word vectors of its class label tokens), but also a class's IRI vector and the concatenation of the word vector and IRI vector, with the best performance reported.
For Onto2Vec and OPA2Vec, we take some different settings as their original papers: the classes (IRIs) in the sentences transformed from axioms are replaced by their labels for the training corpus, and a class's word vector (i.e., the average of the vectors of the label tokens) instead of its IRI vector is adopted for better performance. 

\item \textbf{Geometric knowledge graph (KG) embedding methods.}
Three classic geometric KG embedding methods --- TransE \cite{bordes2013translating}, TransR \cite{lin2015learning} and DistMult \cite{yang2014embedding}, as well as one recent KG embedding method HAKE which takes the class hierarchy into consideration \cite{zhang2020learning}, are used to embed the ontology classes and existential restrictions. 
The embeddings are further fed to an LR or RF classifier for predicting the subsumption's score as the above baselines, or directly used to calculate the subsumption's score according to the embedding method's triple scoring function. The setting that leads to the best validation MRR is used in reporting the final testing results.
It is worth mentioning that the OWL ontologies are transformed into multi-relation graphs following OWL Mapping to RDF Graphs\footnote{\url{https://www.w3.org/TR/owl2-mapping-to-rdf/}}, during which complex classes are represented by blank nodes with some built-in properties and classes (e.g., a property existential restriction $\exists r. C$ is represented by a blank node \_ together with the triples of (\_, \textit{rdf:type}, \textit{owl:Restriction}), (\_, \textit{owl:onProperty}, $r$)  and (\_, \textit{owl:someValuesFrom}, $C$)). 
The annotation properties and their associated literals are filtered out since the literals will be very simply transformed into normal unique entities and their semantics will not be exploited by the above KG embedding methods.

\item \textbf{Text-aware KG embedding methods.}
KG and text joint embedding is also considered for the baseline.
We take the method proposed in \cite{mousselly2018multimodal}, \gyx{
where for each triple, two scores are first computed by the TransE approach using entities' geometric embeddings and text word embeddings, respectively, and then another three scores are computed 
for joint embedding via
}
\textit{(i)} translating the subject's text word embedding to the object's geometric embedding, \textit{(ii)} translating the subject's geometric embedding to the object's text word embedding, and \textit{(iii)} translating the \gyx{sum} of the geometric embedding and text word embedding of the subject to the counterpart of the object. 
We name this method as Text-aware TransE, and run it over the above mentioned KG transformed from the OWL ontology, as well as the entity labels defined by the annotation properties of \textit{rdfs:label} and \textit{obo:hasExactSynonym} (and \textit{obo:hasSynonym}) for GO (FoodOn).
For initial entity text embeddings, we use the above mentioned Word2Vec model.
In the transformed KG, the entities of the original ontology restrictions are blank nodes with no labels; Text-aware TransE will not be able to generate their embeddings, and its performance would be impacted especially on predicting existential subsumptions.
We thus generate labels for the existential restriction in form of $\exists r. C$ using the same approach as in \bertsubs. 
The Text-aware TransE version run on this label augmented KG is denoted as $\text{Text-aware}^{+}\text{ TransE}$.



\end{itemize}

For \bertsubs path context (PC) and breadth-first context (BS), we consider the depth $d$ of $1$ and $2$, respectively, since the too-far-away classes provide little useful information but dramatically increase the average sentence length. For the maximum surrounding subsumptions, we consider the sizes of $2$, $4$ and $6$ for both PC and BC.
The final \bertsubs settings and the baseline settings are adjusted according to their MRR results measured on the validation set.
By default, \bertsubs uses one English label defined by \textit{rdfs:label} for each class. In Section \ref{sec:ablation}, we also report its results using multiple labels defined by some additional annotation properties.
We use Hugging Face transformers to implement \bertsubs\footnote{Our codes and data: \url{https://gitlab.com/chen00217/bert_subsumption}\label{fnref}}, and adopt the pre-trained BERT named ``bert-base-uncased''\footnote{\url{https://huggingface.co/bert-base-uncased}} with the maximum input sequence length set to $128$ for IC and $256$ for PC and BC.  

\subsubsection{Inter-ontology Subsumption Prediction}

We predicted named subsumptions between 
\textit{(i)} NCIT, which is a large ontology composed of various cancer-related concepts including cancer diseases, findings, drugs, anatomy, abnormalities and so on \cite{nicholas2007ncit}, and DOID, which is a regularly maintained ontology about human diseases \cite{Schriml2018do},
and \textit{(ii)} the health lifestyle ontology HeLiS (0.4.8), which contains concepts of food, nutrients and activities \cite{dragoni2018helis}, and the foodon ontology FoodOn.
These two ontology pairs are denoted as NCIT-DOID and HeLiS-FoodOn, respectively.

The ontologies and the inter-ontology subsumptions of NCIT-DOID are from the ontology matching resources constructed in our previous work \cite{he2022machine}.
We first got high quality expert curated equivalence class mappings between NCIT and DOID from the Mondo community\footnote{\url{https://mondo.monarchinitiative.org/}}.
Note since the equivalence mappings are all about classes of diseases, we have pruned the original ontologies of NCIT (V18.05d) and DOID (V1.2) by cutting off classes unrelated to diseases. This reduces the ontology scales and improves the relative completeness of the equivalence mappings.
For each given equivalence mapping between $c_1$ from NCIT and $c_2$ from DOID, we generate a set of subsumption pairs by combing $c_1$ with the declared named class subsumers of $c_2$, denoted as $\left\{(c_1, c_2') | (c_2, c_2') \right\}$. 
Note those subsumers of $c_2$ that are entailed by inheritance reasoning, e.g., the grandparents of $c_2$, are not considered since they are not as fine-grained as the declared subsumers and their prediction is less challenging and less useful.
Meanwhile, as a general evaluation resource, we deleted the original class $c_2$ in DOID to prevent the system from utilizing the equivalence mapping to infer the subsumption. 
Please see \cite{he2022machine} for more details on the construction of the NCIT-DOID data.
For HeLiS-FoodOn, we adopt the original large-scale ontologies and adopted the same method as NCIT-DOID to get subsumptions from $372$ high quality equivalence class mappings that were manually annotated in our previous ontology matching study \cite{chen2021augmenting}.
The statistics of the two inter-ontology subsumption prediction tasks are shown in Table \ref{tab:datasets2}.
The named subsumptions are divided into a validation set ($25\%$) and a testing set ($75\%$). There is no training set because \bertsubs and the baselines use the named subsumptions within each ontology for training. Evaluation under this setting can already verify the performance of \bertsubs. Complementing the training samples by some given inter-ontology subsumptions would improve the performance, but makes the systems less automatic, and is not considered in our current evaluation.

We use the same ranking-based metrics and the same negative subsumer extraction method as intra-ontology named subsumption prediction.
On average, each validation/test subsumption of NCIT-DOID and HeLiS-FoodOn has $38.72$ and $40.7$ negative subsumers, respectively.
We also used the same hyper parameter searching method for \bertsubs, and adopted the none-contextual word embedding-based methods as the baselines. 
By default, \bertsubs also uses one single class label --- the English label defined by \textit{rdfs:label}. For HeLiS, some classes have no label information but their IRI names are meaningful. For these classes, we extract and parse their IRI names (e.g., ``processed legumes'' for \textit{vc:ProcessedLegumes} in Fig. \ref{fig:onto}) as their labels.
We also evaluated using multiple labels.

\begin{table}
\renewcommand\arraystretch{1.3}
\centering
\footnotesize{
\centering
\begin{tabular}{p{2.2cm}<{\centering}|p{2.2cm}<{\centering}|p{2.7cm}<{\centering}}\hline
Ontology Pair & Named Classes & Named Subsumptions  \\ \hline
NCIT - DOID & $6,835$ - $5,113$  & $3,336$    \\ \hline
HeLiS - FoodOn & $20,595$ - $28,308$  & $421$   \\ \hline
\end{tabular}
\caption{
Statistics of the ontology pairs for inter-ontology subsumption prediction. }\label{tab:datasets2}
}
\end{table}

\subsection{Overall Results}

The overall results of intra-ontology named subsumption prediction are shown in Table \ref{res:intra-named} where the best results are bolded.
On the one hand, we can find considering the textual information is beneficial.
On both FoodOn and GO, Word2Vec itself already performs a bit better than the best of the geometric KG embedding methods; Text-aware TransE and $\text{Text-aware}^{+}\text{ TransE}$ both significantly outperform the original TransE; the Word2Vec-based OWL ontology embedding methods, especially \owlvec, all perform quite well.
%
On the other hand, \bertsubs (IC), which only considers the labels of two isolated classes but adopts contextual word embedding with BERT,
further outperforms the OWL ontology embedding methods as well as the other baselines; for example, in comparison with \owlvec, \bertsubs improves MRR from $0.462$ to $0.586$ on GO.
Regarding the templates for utilizing the class contexts, the PC template is quite effective \gyx{in view of the facts that} \bertsubs (PC) outperforms \bertsubs (IC) on MRR and H@$1$ on both FoodOn and GO, while the BC template plays a slight negative impact w.r.t. most metrics.

The overall results on intra-ontology restriction subsumption prediction are shown in Table \ref{res:intra-restriction}.
As intra-ontology named subsumption prediction, the textual information is still quite effective, with much better performance achieved by $\text{Text-aware}^{+}\text{ TransE}$ than the original TransE.
The ontology embedding methods have better performance than the geometric KG embedding methods on GO, but on FoodOn, methods of the latter, including TransR, DistMult and HAKE, in contrast, have better performance, which could be due to that the associated \textit{owl:onProperty} triples and \textit{owl:someValuesFrom} triples can contribute important information in distinguishing the correct and wrong restriction subsumers.
Meanwhile, \bertsubs (IC) dramatically outperforms all the baselines (e.g., the MRR value is improved from $0.757$ to $0.781$ on FoodOn and from $0.700$ to $0.898$ on GO, in comparison with the best baseline), and this improvement is higher than that in intra-ontology named subsumption prediction. 
Both the PC template and the BC template are quite effective w.r.t. the task on both ontologies, in comparison with using isolated classes, and the PC template has slightly better performance than the BC template. 
For example, H@$1$ on FoodOn is improved from $0.670$ to $0.723$ by the PC template, and to $0.706$ by the BC template.

The overall results on inter-ontology named subsumption prediction are shown in Table \ref{res:inter}.
Note the geometric KG embedding baselines cannot not applied in this task since the subsumption is between two independent ontologies. 
The observations on \bertsubs are very close to those of intra-ontology named subsumption prediction: \bertsubs (IC) can already achieve better results than all the baselines (except for H@$1$ on HeLiS-FoodOn in comparison with \owlvec); the PC template is also effective, leading to some performance improvement especially towards the metrics of MRR and H@$1$ on both NCIT-DOID and HeLiS-FoodOn, while the BC template is less effective.

\rv{
It is worth noting \bertsubs (IC) has better performance than \bertsubs (BC) in all the three above scenarios. One potential reason is that the BC template attempts to encode more complete contexts of the two classes. This leads to more complex input sequences that the model needs to learn from, with higher information redundancy and potential noise.
Due to this observation, we do not consider generating more complicated input sequences by using the IC and the BC templates at the same time.
}

%

\begin{table}
\renewcommand\arraystretch{1.3}
\centering
\footnotesize{
\centering
\begin{tabular}{c|cccc|cccc}\hline
\multirow{2}{*}{Method}
& \multicolumn{4}{c|}{FoodOn} &  \multicolumn{4}{c}{GO}\\
& MRR & H@$1$ & H@$5$ & H@$10$ & MRR & H@$1$ & H@$5$& H@$10$\\ \hline
TransE & $0.479$ & $0.332$ & $0.654$ & $0.816$ &$0.320$ &$0.192$ &$0.444$ &$0.605$  \\ \hline
TransR & $0.508$ & $0.367$ & $0.674$ & $0.827$ & $0.354$ &	$0.218$ & $0.497$ & $0.647$   \\ \hline
DistMult & $0.509$ & $0.369$ & $0.678$ & $0.821$ & $0.344$ & $0.216$ & $0.471$ & $0.612$  \\ \hline
HAKE &$0.488$ &$0.349$ &$0.658$ &$0.800$ & $0.416$ &$0.295$ & $0.541$ &$0.654$  \\ \hline\hline
Text-aware TransE & $0.572$ & $0.429$ & $0.734$ &$0.869$ & $0.518$ & $0.357$ & $0.718$ & $0.863$ \\ \hline
$\text{Text-aware}^{+}\text{ TransE}$ & $0.567$ & $0.434$ & $0.730$ & $0.860$ & $0.515$ & $0.354$ & $0.716$ & $0.856$ \\ \hline\hline
Word2Vec & $0.562$ & $0.426$ & $0.717$ & $0.866$  & $0.416$ & $0.284$ & $0.549$ & $0.721$  \\ \hline
Onto2Vec & $0.591$ & $0.451$ & $0.762$ & $0.875$ & $0.428$ & $0.291$ & $0.570$ & $0.751$   \\ \hline
OPA2Vec & $0.607$ & $0.464$ & $0.782$ & $0.892$ & $0.434$ & $0.294$ & $0.585$ & $0.760$   \\ \hline
\owlvec & $0.628$ & $0.502$ & $0.797$ & $0.900$  & $0.462$ & $0.328$ & $0.596$ & $0.787$ \\ \hline\hline
\bertsubs (IC) & $0.635$ & $0.483$ & $\bm{0.832}$ & $0.931$  &  $0.586$ & $0.408$ & $\bm{0.825}$ & $\bm{0.937}$ \\ 
\bertsubs (PC) & $\bm{0.636}$ & $\bm{0.491}$ & $0.829$ & $0.932$ & $\bm{0.606}$ & $\bm{0.453}$ & $0.806$ & $0.927$ \\ 
\bertsubs (BC) & $0.618$ & $0.459$ & $0.824$ & $\bm{0.935}$  &$0.578$ & $0.429$ & $0.767$ & $0.907$ \\ \hline
\end{tabular}
\vspace{-0.2cm}
\caption{
Results of \textbf{intra-ontology name subsumption} prediction. 
}
\label{res:intra-named}
}
\end{table}

\begin{table}
\renewcommand\arraystretch{1.3}
\centering
\footnotesize{
\centering
\begin{tabular}{c|cccc|cccc}\hline
\multirow{2}{*}{Method}
&  \multicolumn{4}{c|}{FoodOn} &  \multicolumn{4}{c}{GO}\\
& MRR & H@$1$ & H@$5$ & H@$10$ & MRR & H@$1$ & H@$5$& H@$10$\\ \hline
TransE  & $0.404$ &$0.284$ &$0.529$ &$0.662$ & $0.133$ & $0.058$&$0.158$ &$0.242$ \\ \hline
TransR  &$0.713$ &$0.633$ &$0.810$ & $0.869$& $0.462$ &$0.337$ &$0.591$ &$0.712$  \\ \hline
DistMult  &$0.757$ &$0.696$ &$0.821$ &$0.868$ &$0.452$ &$0.279$ &$0.684$ & $0.834$ \\ \hline
HAKE & $0.700$ &$0.633$ &$0.768$ &$0.808$ & $0.564$ & $0.512$ & $0.604$ & $0.643$\\ \hline\hline
$\text{Text-aware}^{+}\text{ TransE}$ & $0.622$ & $0.514$ & $0.752$ &$0.834$ & $0.615$ & $0.490$  &$0.772$ &$0.877$ \\ \hline\hline
Word2Vec&$0.568$ &$0.459$ & $0.691$&$0.777$ &$0.588$ & $0.485$ & $0.697$ & $0.809$ \\ \hline
Onto2Vec  &$0.650$ & $0.550$& $0.770$& $0.846$&$0.680$ &$0.568$ &$0.818$ &$0.907$  \\ \hline
OPA2Vec  &$0.629$ &$0.528$ &$0.743$ &$0.826$ & $0.689$ &$0.584$ &$0.818$ &$0.908$  \\ \hline
\owlvec & $0.654$& $0.562$&$0.757$&$0.826$&$0.700$ &$0.593$ &$0.833$ & $0.915$ \\ \hline\hline
\bertsubs (IC)  & $0.781$ &$0.679$ &$0.919$ &$0.947$ &$0.898$ &$0.850$ &$0.958$ &$0.975$ \\ 
\bertsubs (PC) &$\bm{0.814}$ &$\bm{0.723}$ &$\bm{0.932}$ &$\bm{0.963}$ &$\bm{0.917}$ &$\bm{0.878}$ &$\bm{0.966}$ &$\bm{0.985}$ \\ 
\bertsubs (BC) &$0.807$ &$0.706$ &$\bm{0.932}$ &$0.962$ & $0.915$ &$0.876$ &$0.962$ &$0.981$  \\ \hline
\end{tabular}
\vspace{-0.2cm}
\caption{
Results of \textbf{intra-ontology restriction subsumption} prediction. 
}\label{res:intra-restriction}
}
\end{table}

\begin{table}
\renewcommand\arraystretch{1.3}
\centering
\footnotesize{
\centering
\begin{tabular}{c|cccc|cccc}\hline
\multirow{2}{*}{Method}
&  \multicolumn{4}{c|}{NCIT-DOID} &  \multicolumn{4}{c}{HeLiS-FoodOn}\\
& MRR & H@$1$ & H@$5$ & H@$10$ & MRR & H@$1$ & H@$5$& H@$10$\\ \hline
Word2Vec  & $0.444$ &$0.320$ &$0.575$ &$0.722$ & $0.541$ &$0.415$ &$0.712$ &$0.810$ \\ \hline
Onto2Vec  & $0.485$ &$0.351$ &$0.637$ &$0.784$ & $0.592$ & $0.465$&$0.725$ &$0.842$ \\ \hline
OPA2Vec & $0.488$ &$0.367$ &$0.641$ &$0.784$ & $0.588$  &$0.449$ &$0.731$ &$0.870$ \\ \hline
\owlvec & $0.506$ &$0.378$ &$0.650$ &$0.784$ & $0.610$ &$\bm{0.501}$ &$0.753$ &$0.839$ \\ \hline\hline
\bertsubs (IC)  & $0.695$ &$0.574$ &$0.854$ &$\bm{0.935}$ & $0.619$ & $0.449$ & $\bm{0.858}$ &$\bm{0.936}$ \\ \hline
\bertsubs (PC) & $\bm{0.707}$ &$\bm{0.588}$ &$\bm{0.863}$ &$0.934$ &$\bm{0.629}$ &$0.481$ &$0.842$ &$0.927$ \\ \hline
\bertsubs (BC) & $0.693$ &$0.565$ &$0.851$ &$0.929$ &$0.589$ &$0.440$ & $0.767$& $0.875$\\ \hline
\end{tabular}
\vspace{-0.2cm}
\caption{
Results of \textbf{inter-ontology named subsumption} prediction. 
}\label{res:inter}
}
\end{table}

\subsection{Ablation Studies}\label{sec:ablation}

\subsubsection{Context Hops and Maximum Subsumptions}
\begin{figure}
\centering
\includegraphics[width=0.95\textwidth]{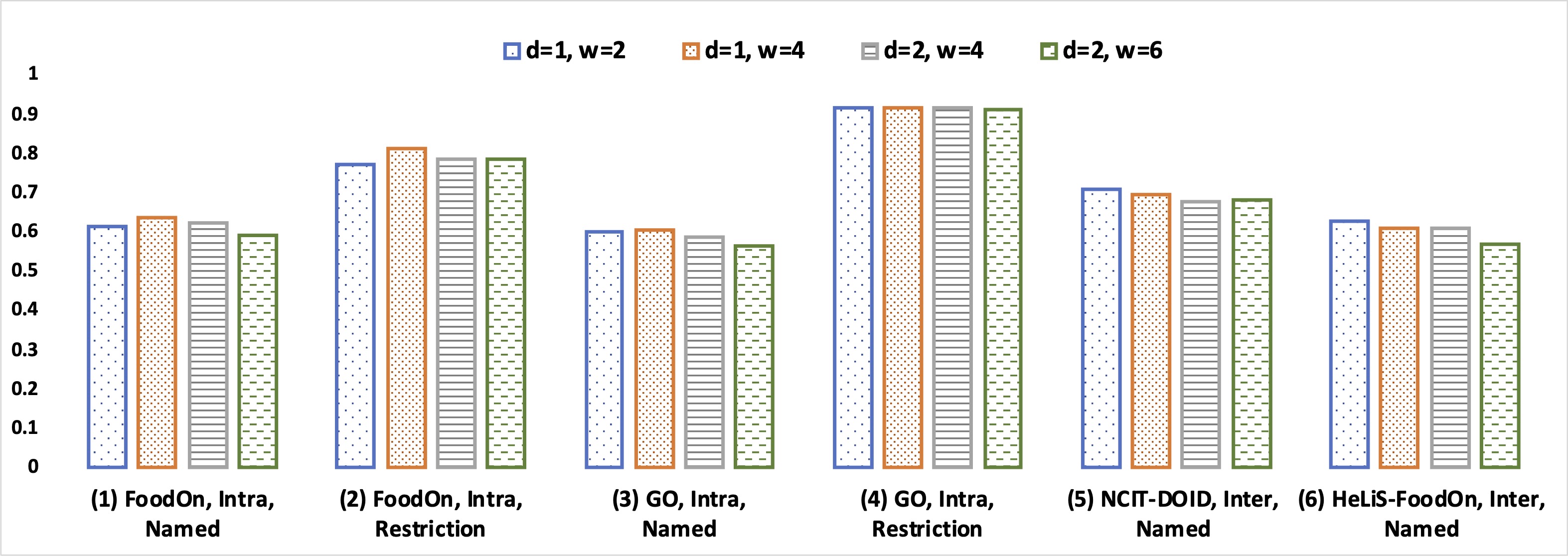}
\caption{MRR results of \bertsubs (PC) under different settings of the context hops ($d$) and the maximum subsumptions ($w$)} \label{fig:pc}
\end{figure}

The overall result analysis has \gyx{presented} the effectiveness of the PC template and the BC template of \bertsubs.
The purpose of this part is to analyze the impact of different settings towards the context in the class hierarchy.
In Fig. \ref{fig:pc}, we show the MRR results of \bertsubs (PC) with different combinations of the context hops ($d$) and the maximum subsumptions ($w$) on all the subsumption prediction tasks with all the ontologies.
We \gyx{opt to report} the results of \bertsubs (PC) for analysis in consideration of its higher overall performance on all the tasks.
We can find that the setting of ($d$ = $1$, $w$ = $4$) has higher MRR values than all the three other settings on the tasks of (1), (2) and (3), while on the task of (4), the setting of ($d$ = $1$, $w$ = $4$) also performs quite well, being very close to the best.
Meanwhile, the setting of ($d$ = $1$, $w$ = $2$), which only considers two directly connected classes, also performs quite well on most tasks, especially (5) and (6) for inter-ontology subsumption prediction.
In contrast, considering the surrounding classes within two hops ($d$ = $2$) does not perform as well as just considering one-hop surrounding classes ($d$ = $1$), except on the task of (4) where all four settings perform very closely.
This is also the reason why we do not test \bertsubs with surrounding classes beyond 2 hops.

\subsubsection{Single Label vs Multiple Labels}\label{sec:label}
Since the textual information plays an important role in subsumption prediction, we compared using multiple class labels with using one single class label in \bertsubs, using the IC and PC templates.
In our current \bertsubs implementation, the solution of using multiple labels is still simple: it replaces a class \rv{by} multiple labels, leading to
multiple samples for one subsumption.
In training, it improves the sample size, while in testing, the scores of multiple samples are averaged.
Besides \textit{rdfs:label},  we adopt the following additional annotation properties for multiple class labels:  \textit{obo:hasSynonym} and \textit{obo:hasExactSynonym} for FoodOn with $1.7$ labels per class, \textit{obo:hasExactSynonym} for GO with $2.9$ labels per class, 
\textit{ncit:P90}\footnote{\textit{ncit:} denotes the prefix of \url{http://ncicb.nci.nih.gov/xml/owl/EVS/Thesaurus.owl\#}}
(``fully qualified synonym'') and \textit{ncit:P108} (``preferred name'') for NCIT with $3.6$ labels per class, \textit{obo:hasExactSynonym} for DOID with $1.9$ labels per class.
HeLiS has no additional labels besides the one by \textit{rdfs:label} (or the IRI name).
The results are shown in Table \ref{res:labels}.
We can find that using multiple labels leads to a bit higher performance in three cases (i.e., intra-ontology named subsumption prediction on FoodOn, intra-ontology restriction subsumption prediction on FoodOn and GO), while in the remaining nine cases, it brings no improvement or even leads to worse performance, especially for \bertsubs (PC) where the labels of not only the target class but also the contextual classes are combined and fed into the model.    
One potential reason is that labels by many additional annotation properties actually have lower quality than the label by \textit{rdfs:label}.
Meanwhile, different annotation properties would have a bit different meanings, and thus simply regarding them as the same would lead to inconsistency in fine-tuning the BERT model.

\begin{table}
\renewcommand\arraystretch{1.3}
\centering
\footnotesize{
\centering
\begin{tabular}{p{0.74cm}<{\centering}|p{0.45cm}<{\centering}p{0.55cm}<{\centering}|p{0.45cm}<{\centering}p{0.55cm}<{\centering}|p{0.45cm}<{\centering}p{0.55cm}<{\centering}|p{0.45cm}<{\centering}p{0.55cm}<{\centering}|p{0.45cm}<{\centering}p{0.55cm}<{\centering}|p{0.45cm}<{\centering}p{0.4cm}<{\centering}}\hline
 \multirow{3}{*}{Setting} &  \multicolumn{4}{c|}{Intra, Named} &  \multicolumn{4}{c|}{Intra, Restriction} &\multicolumn{4}{c}{Inter, Named}\\
  & \multicolumn{2}{c|}{FoodOn} &  \multicolumn{2}{c|}{GO} &  \multicolumn{2}{c|}{FoodOn} &  \multicolumn{2}{c|}{GO} &  \multicolumn{2}{c|}{NCIT-DOID} &  \multicolumn{2}{c}{HeLiS-FoodOn}\\ 
& MRR &H@$5$ & MRR &H@$5$ & MRR & H@$5$ & MRR & H@$5$ & MRR & H@$5$ & MRR & H@$5$ \\ \hline
IC, S  & $0.635$ & $0.832$ &$\textbf{0.585}$ & $\textbf{0.825}$ &$0.781$& $0.919$ &$0.898$& $0.958$ & $\textbf{0.695}$& $\textbf{0.854}$ & $\textbf{0.619}$ & $\textbf{0.858}$  \\ 
IC, M  & $\textbf{0.639}$& $\textbf{0.836}$ &$0.531$ & $0.777$ &$\textbf{0.784}$& $\textbf{0.942}$ &$\textbf{0.899}$ & $\textbf{0.959}$& $0.672$ & $0.847$ &$0.613$ & $0.823$ \\ \hline\hline
PC, S & $\textbf{0.636}$& $\textbf{0.829}$ &$\textbf{0.606}$ & $0.806$ &$\textbf{0.814}$&  $0.932$ & $\textbf{0.917}$& $\textbf{0.966}$ & $\textbf{0.707}$ & $\textbf{0.863}$ & $\textbf{0.629}$ & $\textbf{0.842}$  \\ 
PC, M  & $0.611$& $0.826$ &$0.584$ & $\textbf{0.811}$ &$0.807$& $\textbf{0.933}$ &$0.892$ & $0.955$ & $0.641$ &$0.840$  &$0.574$ &$0.769$    \\ \hline
\end{tabular}
\vspace{-0.2cm}
\caption{
Results of \bertsubs (\textbf{IC}) and (\textbf{PC}), using single \textbf{(S)} label vs multiple \textbf{(M)} labels 
}\label{res:labels}
}
\end{table}

\section{Related Work}\label{sec:related}

\subsection{Ontology Embedding}
In the past decade, quite a few methods have been developed for embedding and completing knowledge graphs (KGs) that are composed of relational facts \cite{wang2017knowledge}.
These methods such as TransE \cite{bordes2013translating} and TransR \cite{lin2015learning} can be applied to predict subsumption relationships in OWL ontologies by transforming the ontologies into RDF graphs (triple sets). 
Besides the W3C OWL Mapping to RDF Graphs we used in evaluation, some other transformations, e.g., the projection rules used on ontology visualization \cite{soylu2018optiquevqs}, could be considered, but these transformations often miss some semantics in OWL, especially the logical expressions.  
Meanwhile, some geometric KG embedding methods, such as HAKE \cite{zhang2020learning}, Poincar\'{e} Embedding \cite{nickel2017poincare} and Box Lattice Embeddings \cite{lees2020embedding}, have been proposed to consider the hierarchical structure.
But they do not consider other semantics beyond the graph structure in OWL ontologies, making them fail to utilize these semantics especially the literals and the logical expressions.
Some literal-aware KG embedding methods, such as Text-aware TransE \cite{mousselly2018multimodal} evaluated in our study, have also been proposed to jointly embed the graph and the literals that are associated by some data properties \cite{gesese2019survey}, but they usually aim at relational facts instead of OWL ontologies with little attention to e.g., the hierarchical classes.

Recently, the geometric KG embedding methods have been further extended to embed logical relationships in OWL ontologies with more complex modeling in the vector space (e.g., classes are represented by areas instead of points). Typical methods include EL Embedding \cite{kulmanov2019embeddings}, Box EL Embedding \cite{xiong2022box} and Logic Tensor Networks \cite{ebrahimi2021towards} for Description Logic $\mathcal{EL}^{++}$, and Quantum Embedding \cite{garg2019quantum} for Description Logic $\mathcal{ALC}$.
As those classic geometric KG embeddings, these methods currently only consider (a part of) the logical relationships in an OWL ontology.
On the other hand, some recent OWL ontology embedding methods such as OWL2Vec* \cite{chen2021owl2vec}, OPA2Vec \cite{smaili2019opa2vec} and Onto2Vec \cite{smaili2018onto2vec}, which first transform the ontology contents into entity and/or word sequences with entity and word correlation kept, and then learn word embedding models to get word and entity vectors, have achieved quite good performance on ontology relevant prediction tasks such as class subsumption prediction, class clustering and gene–disease association prediction.
Their weakness mainly lies in the adopted non-contextual word embedding model which has been proven to perform worse than the more recent Transformer-based contextual word embedding models such as BERT in many natural language understanding and sequence learning tasks.

\subsection{Pre-trained Language Model and Knowledge Graph}

Transformer-based PLMs such as BERT and T5 have achieved great success in many taks of natural language inference such as text classification, question answering and machine reading comprehension \cite{devlin2019bert,lee2020biobert,raffel2020exploring}.
With the recent development of prompt techniques, these PLMs can be further fine-tuned and applied to different kinds of tasks, where input with different contexts and formats are transformed, embedded and exploited \cite{liu2021pre}. 
There have been some works that apply PLMs to address KG completion and some other prediction tasks of KGs.
For example, KG-BERT \cite{yao2019kg} fine-tunes a BERT to predict a relational fact (triple) by encoding its text (i.e., the labels of its entities and relation); similarly, COMET \cite{bosselut2019comet} uses a pre-trained Transformer Language Model to predict the fact of ConceptNet by its text; Wang et al. \cite{wang2021structure} follow the text encoding paradigm of KG-BERT by augmenting the text embedding with the graph embedding so as to take the entity's context into consideration.
There are also some studies that utilize the symbolic knowledge in a KG as the reference to probe the sub-symbolic knowledge in PLMs.
One typical work is LAMA \cite{petroni2019language} which uses the pre-trained BERT to predict the masked tokens of natural language sentences, each of which is generated from one KG fact by a simple template.
These works are quite different from \bertsubs which aims at OWL ontologies that have more complex knowledge formats than relational facts and require different templates.
\bertsubs can also support both inter-ontology and intra-ontology subsumptions, as well as the existential restrictions as the subsumers. All these tasks are quite different from the related works presented above.

BERT has been applied to ontology subsumption prediction and ontology matching, but the research of this direction is still in an early stage.  
Liu et al. \cite{liu2020concept} predict subsumers in the SNOMED ontology for new and isolated classes to insert, where templates are used to generate sentences for the class context as BERT input. In contrast, \bertsubs aims at the subsumption between two existing classes, both of which have contexts, with different templates and fine-tuning mechanisms, and more importantly considers the property existential restriction as the subsumer.
Our previous work \bertmap \cite{he2022bertmap} predicts the equivalence mapping (i.e., equivalent classes) between two OWL ontologies, using the labels of the two classes as the IC template introduced in this paper.
\bertsubs is a follow-up work of \bertmap with new research on different ontology curation tasks, more complicated templates for utilizing the class context, and the support of existential restrictions.

\section{Conclusion and Discussion}\label{sec:discussion}
In this paper we present a new method named \bertsubs which fine-tunes the pre-trained language model BERT for predicting class subsumptions of OWL ontologies, with three kinds of templates --- Isolated Class (IC), Path Context (PC), and Breadth-first Context (BC) developed to exploit the text information and the surrounding classes of the target classes. 
It can predict subsumers of not only named classes and property existential restrictions within an ontology for ontology completion, but also named classes in a different ontology for ontology subsumption matching.
Extensive evaluation has been conducted on all these three tasks with $6$ real-world ontologies.
\bertsubs often dramatically outperforms all the baselines that are based on (literal-aware) geometric KG embedding methods and the state-of-the-art OWL ontology embedding methods.
Meanwhile, we verify that the the PC template, which takes the path alike context into consideration, often performs better than the simple IC template, while the BC template is not always effective.


The subsumers in OWL ontologies often contain more than basic property existential restrictions in form of $\exists r. C$. 
On the one hand, they can be the composition of some such basic restrictions and some other named classes via conjunction, disjunction and/or negation operations (e.g., $\neg D \sqcap \exists r. C$). We can use \bertsubs to predict such subsumers by first predicting its components independently, and then combining their scores according to the logical operations. However such a simple solution cannot consider the relationship between these components.   
On the other hand, the restriction class $C$ itself could be a complex class composed of some named classes and other existential restrictions.
Meanwhile, although the property existential restriction is the most widely used restriction in many ontologies due to its light reasoning cost, other restrictions such as the universal restriction in form of $\forall r.C$ and the at-least restriction in form of $\ge n r.C$ (which restricts that at least $n$ individuals that can be reached via the relation $r$) are often used in OWL ontologies.
 To support all these complex subsumers, some more complicated templates need to be manually developed, or automatically generated (learned) via a general approach.
Some previous works on generating natural language descriptions for OWL ontologies (a.k.a. OWL verbalization) \cite{stevens2011automating,kaljurand2007attempto} could be considered for creating the templates, but the their generated text may not exactly match the prediction task, since the purpose of these works, which are to provide ontology interfaces for human beings, is different from \bertsubs.

Besides the promising performance of \bertsubs, the evaluation also shows that the current simple solution of utilizing multiple labels by different annotation properties still has ample space for further improvement.
Exploiting the class context besides surrounding classes, such as different annotation properties, data properties, and logical expressions, is still a big challenge for the further research, for not only improving \bertmap but also other BERT applications in OWL ontologies.
Meanwhile, we currently use the class hierarchy extracted from just declared subsumptions for computing the class context. 
In the future evaluation, the class context computed by entailment reasoning will also be considered.
In the end, we will further maintain and \gyx{optimize} the codes of \bertsubs to improve its usability, \rv{evaluate it with more ontologies in different domains to show its generality}, and make it as a part of our DeepOnto\footnote{\url{https://github.com/KRR-Oxford/DeepOnto}\label{fndeeponto}} library (which is to support ontology construction and curation using PLMs, machine learning and semantic techniques) together with our ontology matching method \bertmap \cite{he2022bertmap}.

\section{Declarations}

\noindent\textbf{Ethics Approval}.
Not applicable. The paper does not involve any ethics issues.

\noindent\textbf{Conflicts of Interest/Competing Interests}.
This paper has conflicts of interest with members from The University of Manchester (manchester.ac.uk), University of Oxford (ox.ac.uk), Zhejiang University (zju.edu.cn), City University of London (city.ac.uk) and University of Oslo (uio.no).

\noindent\textbf{Authors' Contributions}.
Jiaoyan Chen is the corresponding author, with major contribution to the paper discussion, experiments and writing. His work was mainly done when he was in University of Oxford. Yuan He and Yuxia Geng contributed partially to the paper discussion and experiments. Ernesto Jimenez-Ruiz, Hang Dong and Ian Horrocks contributed partially to the paper discussion and writing.



\noindent\textbf{Funding}.
This research was funded in whole or in part by the SIRIUS Centre for Scalable Data Access (Research Council of Norway, project 237889), eBay, Samsung Research UK, Siemens AG, and the EPSRC projects OASIS (EP/S032347/1), UK FIRES (EP/S019111/1) and ConCur (EP/V050869/1). 
For the purpose of Open Access, the author has applied a CC BY public copyright licence to any Author Accepted Manuscript (AAM) version arising from this submission

\noindent\textbf{Availability of Data and Material}.
All the data used in the experiments have been open. The materials used in the paper, such as the figures, will also be open.
%
The codes used in the experiments have been open. They will be further maintained and kept open as a part of the DeepOnto library.

%
%
\bibliographystyle{splncs04}
\bibliography{reference}

\end{document}